\documentclass[final]{cvpr}

\usepackage{times}
\usepackage{epsfig}
\usepackage{graphicx}
\usepackage{amsmath}
\usepackage{amssymb}
\usepackage{colortbl}
\usepackage[ruled,vlined]{algorithm2e}
\usepackage{algorithmic}
\usepackage{subfigure}
\usepackage{breqn}
\usepackage{dsfont}

\usepackage[pagebackref=true,breaklinks=true,colorlinks,bookmarks=false]{hyperref}

\def\cvprPaperID{5287} 
\def\confYear{CVPR 2021}

\begin{document}

\title{Consistent Instance False Positive Improves Fairness in Face Recognition}

\author{Xingkun Xu$^{\dag}$ ~ ~ Yuge Huang$^{\dag}$ ~ ~ Pengcheng Shen$^{\dag}$\thanks{denotes the corresponding author.} ~ ~ Shaoxin Li$^{\dag}$ \\
~ ~ Jilin Li$^{\dag}$  ~ ~ Feiyue Huang$^{\dag}$ ~ ~ Yong Li$^{\S}$ ~ ~ Zhen Cui$^{\S}$ \\
$^\dag$Youtu Lab, Tencent  ~ ~ ~  $^\S$Nanjing University of Science and Technology\\
{\tt\small $^\dag$\{xingkunxu, yugehuang, quantshen, darwinli, jerolinli, garyhuang\}@tencent.com} \\
{\tt\small $^\S$\{yong.li, zhen.cui\}@njust.edu.cn}\\
}

\maketitle

\begin{abstract}
Demographic bias is a significant challenge in practical face recognition systems. Existing methods heavily rely on accurate demographic annotations. However, such annotations are usually unavailable in real scenarios.
Moreover, these methods are typically designed for a specific demographic group and are not general enough.
In this paper, we propose a false positive rate penalty loss, which mitigates face recognition bias by increasing the consistency of instance False Positive Rate (FPR).
Specifically, we first define the instance FPR as the ratio between the number of the non-target similarities above a unified threshold and the total number of the non-target similarities. The unified threshold is estimated for a given total FPR.  Then, an additional penalty term, which is in proportion to the ratio of instance FPR overall FPR, is introduced into the denominator of the softmax-based loss. The larger the instance FPR, the larger the penalty.  By such unequal penalties, the instance FPRs are supposed to be consistent.  Compared with the previous debiasing methods, our method requires no demographic annotations. Thus, it can mitigate the bias among demographic groups divided by various attributes, and these attributes are not needed to be previously predefined during training.  Extensive experimental results on popular benchmarks demonstrate the superiority of our method over state-of-the-art competitors.
Code and pre-trained models are available at \url{https://github.com/Tencent/TFace}.
\end{abstract}

\section{Introduction}
With the increasing deployment of face recognition systems, fairness in face recognition has received broad interest from research communities~\cite{wang2019racial, drozdowski2020demographic, gong2020mitigating,sixta2020fairface, robinson2020face, yucer2020exploring}. 
This is partially due to the enormous impact brought in our daily life by face recognition systems.
For example, when automatic face recognition is applied to crime prevention, unfair prediction may lead to unfair treatment of individuals across different demographic groups.

\begin{figure}[t]
  \centering
  \includegraphics[trim={0 0 0 0mm},clip,width=1\linewidth]{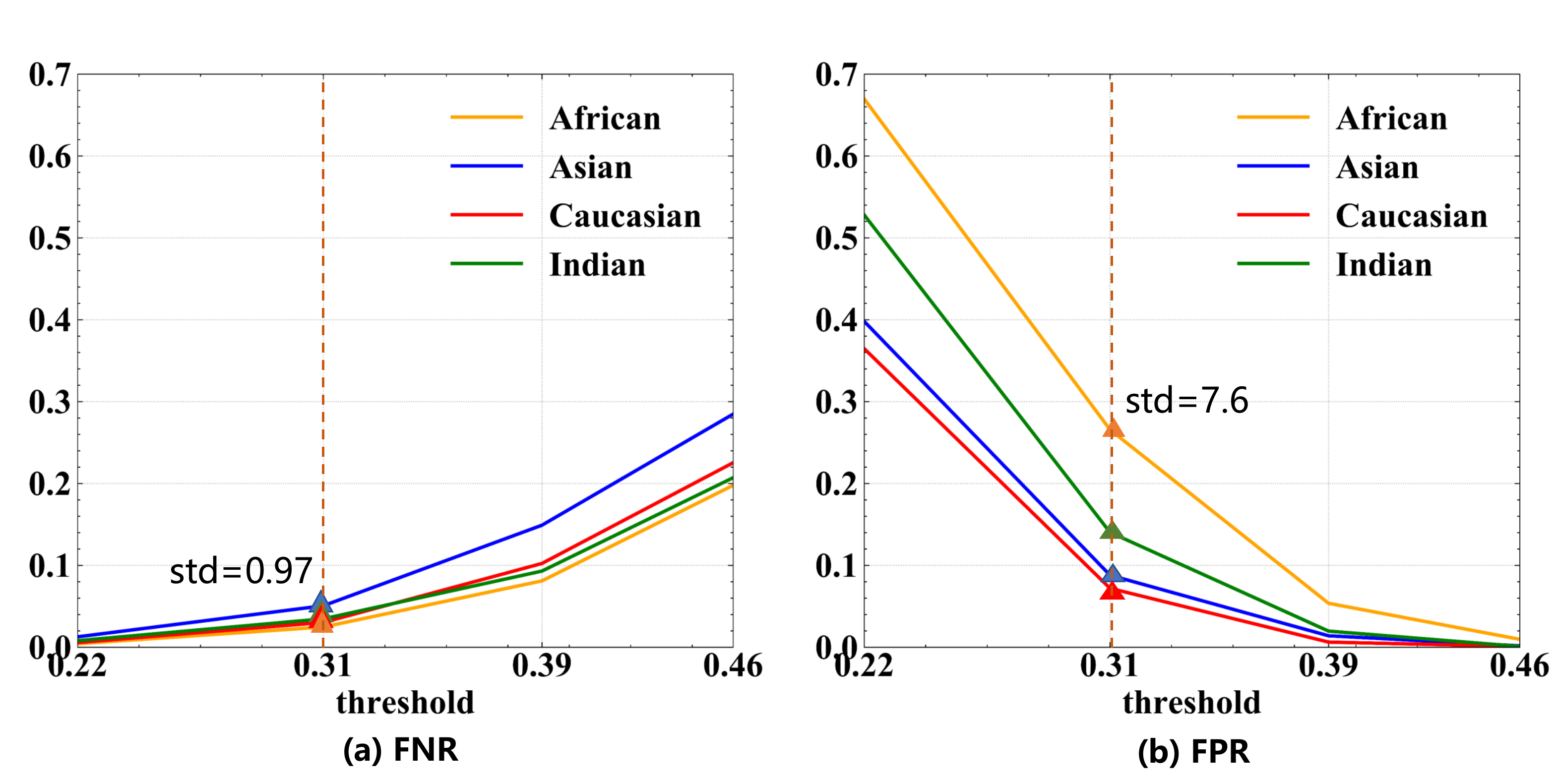}
  \caption{\small FNR and FPR curves of the four races in RFW~\cite{wang2019racial}. 
  FNR and FPR are calculated with a ResNet34~\cite{deng2019arcface}, which is trained on the public balanced dataset BUPT-Balanced~\cite{wang2019mis}.
  Lower is better. Given a specific threshold, PPR varies significantly among different races than FNR (\textit{e.g.}, the standard deviation (std) of FPR at $T_{u}$=$0.31$ is $7.6$, while the std of FNR at $T_{u}$=$0.31$ is $0.97$).}
  \label{fig:demographic_difference_in_RFW}
\end{figure}

Previous studies~\cite{robinson2020face, wang2019racial, drozdowski2020demographic, gong2020jointly, srinivas2019face} mainly improve the fairness of face recognition in two aspects, \textit{i.e.}, datasets and algorithms. 
Since the widely-used public large-scale face datasets, such as CASIA-WebFace~\cite{Yi2014learning}, VGGFace2~\cite{cao2018vggface2}, and MS-Celeb-1M~\cite{guo2016ms1m} are collected from the Internet, they inevitably encode gender, ethnic, and culture biases.
Thus, the works in ~\cite{robinson2020face, wang2019racial, wang2020mitigating, hupont2019demogpairs} propose some new face recognition datasets that contain relatively balanced samples in ethnicity, age, and other facial attributes.
However, it is quite challenging to construct a balanced dataset in various attributes.
What is more, the racial bias of the models trained with such balanced datasets cannot be eliminated completely~\cite{wang2020mitigating}. 
Therefore, a novel algorithm that can mitigate the bias regardless of whether training datasets are balanced or not is imperative.
Recently, several algorithms supervised by demographic attribute information are introduced to alleviate demographic bias.  
For example, Wang et al.~\cite{wang2019racial} propose a deep information maximization adaptation network by transferring recognition knowledge from Caucasians to other races. 
With similar ideas, they propose another method based on a widely-used margin-based loss function in face recognition, in which Q-learning learns the optimal margins of non-Caucasians with a manually-selected margin of Caucasians~\cite{wang2020mitigating}. 
Different from the above methods take Caucasians as a reference, Gong et al.~\cite{gong2020jointly} present a debiasing adversarial network with four specific classifiers, in which one classifier is designed for identity and the other three are designed for demographic attributes.
They further introduce a group adaptive classifier by using adaptive convolution kernels and attention mechanisms based on their demographic attributes~\cite{gong2020mitigating}.
However, all the above methods are explicitly designed to mitigate the bias in demographic groups divided by race. Thus, these methods have poor transferability and generalization.
Moreover, they rely on accurate demographic attribute annotations, which are usually not available.

To address the above problem, we first evaluate the bias in face recognition from another perspective.
Previous methods~\cite{gong2020jointly, wang2020mitigating, wang2019racial} mainly adopt the standard deviation of accuracy in each demographic group as the bias of a specific face recognition algorithm.
In contrast, we analyze the bias in face recognition by two commonly-used evaluation metrics, \textit{i.e.}, false positive rate (FPR), and false negative rate (FNR). 
As shown in Fig.~\ref{fig:demographic_difference_in_RFW}, FPR varies significantly among different races than FNR, which shares a similar observation with~\cite{ngan2015face}. Thus, it is essential to promote the consistency of FPR across each race group to mitigate the bias in face recognition.
Based on this observation, we propose a false positive rate penalty loss, which mitigates face recognition bias by increasing the consistency of instance FPR. 
By generalizing the consistency of FPRs across each demographic group to the consistency of FPRs across each instance, our method is generic to improve the fairness of face recognition across the demographic groups divided by various attributes, such as race, gender, and age.
Specifically, we first define the instance FPR as the ratio between the number of the non-target similarities above a unified threshold and the total number of the non-target similarities.
Then, an additional penalty term in proportion to the ratio of instance FPR overall FPR is introduced into the denominator of the softmax-based loss.
A larger ratio between each instance and the overall FPR yields a larger loss value. 
By such unequal penalties, the instance FPRs are supposed to be much consistent.
Compared with the previous debiasing methods, our method firstly requires no demographic annotations of images; secondly can be easily embedded into the commonly used softmax-based loss function in face recognition; and finally can mitigate the bias across all demographic group divided by various kinds attributes, such as race, gender, and age.

To sum up, the contributions of this work are three-fold:
\begin{itemize}
\item To our best knowledge, it is the first work that alleviates the bias in face recognition by promoting the consistency of instance FPRs, which provides a new perspective to improve face recognition fairness.
\item Our false positive rate penalty loss can improve the fairness across demographic groups divided by various kinds of attributes. Moreover, our method requires no demographic group annotation.
\item We conduct extensive experiments on popular facial benchmarks, which demonstrate the superiority of our method over the SOTA competitors.
\end{itemize}

\begin{figure*}[t!]
  \centering
  \includegraphics[trim={0 0 0 0mm},clip,width=0.97\linewidth]{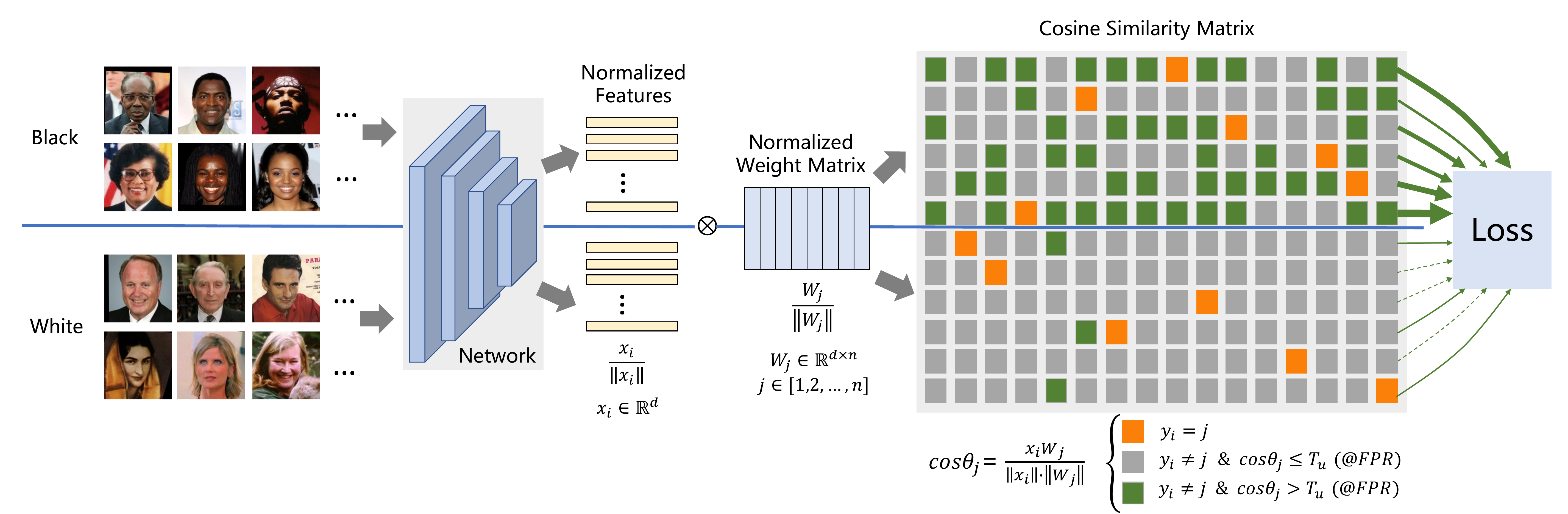}
  \caption{\small Illustration of instance FPR Penalty Loss.
  Assuming a mini-batch input $X$ consists of samples from two races, \textit{i.e.,} black and white, we obtain the corresponding features by an embedding network. 
  Given a weight matrix $W$ that each column corresponds to one identity, a cosine similarity matrix $S$ is calculated based on the normalized feature $X$ and weight $W$, and each value in this matrix is $S_{ij}$=$X_{i}W_{j}$.
  Among the matrix, the orange boxes the cosine similarities between the samples and their corresponding target (ground truth) weights. The other boxes denote the cosine similarities between the samples and the non-target weights. The green boxes indicate that their similarities are above a unified threshold $T_{u}$ estimated by a preset FPR, while the grey boxes indicate equal to or less than the threshold.
  We take the green boxes in each row as false positives and define the instance FPR as the ratio between the number of the green boxes and the total number of the gray and green boxes. For samples from different races, the instance FPR varies significantly.
  Generally, the larger the instance FPR, the worse an algorithm performs at the current training stage. Thus, we introduce an additional false positive penalty term into the softmax-based losses to promote the consistency of instance FPRs.
  }
  \label{fig:framework}
\end{figure*}

\section{Related Work}
\paragraph{Loss Function.} 
Designing an effective loss function plays a vital role in deep face recognition.
Many margin-based loss functions are proposed to obtain highly discriminative features for face recognition.
For example, SphereFace~\cite{liu2017sphereface}, CosFace~\cite{wang2018cosface}, ArcFace ~\cite{deng2019arcface} are widely used margin-based loss function, which add margins in the positive logits (\textit{i.e.}, intra-class).
Recently, several works~\cite{wang2019mis, huang2020curricularface} extend the margin-based loss function with hard sample mining strategies, which add the extra margin in the negative logits (\textit{i.e.}, inter-class). 
Though the loss mentioned above functions are verified to obtain good performance, they do not consider the demographic bias.
Our method can mitigate the bias in face recognition by promoting the consistency of instance FPRs, and thus improve face recognition fairness.
\paragraph{Bias Mitigation in Face Recognition.}
Firstly, we investigate the current datasets widely used for fair face recognition. 
The Diversity in Faces (DiF) dataset~\cite{merler2019diversity} provides annotations of $1$ million human facial images to advance the study of fairness in facial recognition.
Wang et al.~\cite{wang2020mitigating} propose the Racial faces in-the-wild (RFW) as a testing database for studying racial bias in face recognition.
In~\cite{wang2019mis}, Wang et al. also introduce BUPT-balanced as a balanced dataset on race, and BUPT-Globalface to reveal the real distribution of the world’s population. Both of these two public datasets are used for face recognition fairness studies.
In~\cite{robinson2020face}, the BFW benchmark and dataset, inspired by DemogPairs~\cite{hupont2019demogpairs}, is introduced as a labeled data resource made available for evaluating recognition systems. 
BFW contains eight demographic groups for bias evaluation, and each of them consists of $200$ subjects with $2.5$K images.
In this paper, we will evaluate our proposed algorithms on RFW and BFW. 
Next, we discuss the current algorithms for mitigating bias, which aims to solve unfairness of discrimination performance across groups, based on demographic information. 
Wang et al.~\cite{wang2019racial} propose a deep information maximization adaptation network to alleviate bias, using deep unsupervised domain adaptation. 
Subsequently, Wang et al.~\cite{wang2020mitigating} introduce a reinforcement learning based race balance network, in which additive angular margin of loss functions for different races is selected by a pre-trained network module.
Gong et al.~\cite{gong2020jointly} present a debiasing adversarial network with four specific classifiers, in which one classifier for identity and the other three for demographic attributes.
They have further improved the method with a group adaptive classifier based on estimated demographic attributes recently~\cite{gong2020mitigating}.

Manual annotations of demographic attribute are necessary in current studies, which are usually unavailable in practice. Auxiliary modules, such as DQN, MDP, and attribute classifier, increase the training pipelines' difficulty than standard end-to-end training methods. In contrast, our proposed approach is simple to implement in an end-to-end manner without accurate manual annotations and auxiliary network modules.

\section{Proposed Approach}
In this section, we introduce the details of our approach. First, we explain the relationship between false positive rate and bias in face recognition; then we deduce a new evaluation protocol for demographic bias from the corresponding FPRs. Next, we introduce our false positive rate penalty loss, which mitigates face recognition bias by increasing the consistency of instance FPR.
\subsection{Demographic Bias}
\paragraph{False Positive Rate vs. Bias}
In face recognition systems, a comparison of two images with the same identity generates a positive pair, while a comparison of two images with different identities generates a negative pair.
In general, a unified threshold $T_{u}$ should be set as the criterion for judging whether a comparison of two images is positive or negative.
A negative pair with similarity above the threshold is called a false positive pair (FPP), and a positive pair with similarity below the threshold is called a false negative pair (FNP).
Correspondingly, the false positive rate (FPR) is defined as the ratio of false positive pairs to all negative pairs, and the false negative rage (FNR) is defined as the ratio of  false negative pairs to all positive pairs.
Both FPR and FNR are the frequently used as evaluation protocols in face recognition.
Given a similarity set of $N^{+}$ positive pairs ${\{S^{+}[i]\}}$, and a similarity set of $N^{-}$ negative pairs ${\{S^{-}[i]\}}$,
FPR and FNR, which are respectively denoted as $\gamma^{+}$ and $\gamma^{-}$, are formulated as follows:
\begin{equation}
\label{eq:false_positive_rate}
\gamma^{+}=\frac{\sum_{i=1}^{N^{-}} \mathds{1}({S}^{-}[i] > T_{u})}{N^{-}} ,
\end{equation}
\begin{equation}
\label{eq:false_negative_rate}
\gamma^{-}=\frac{\sum_{i=1}^{N^{+}} \mathds{1}({S}^{+}[i] < T_{u})}{N^{+}} ,
\end{equation}
where $\mathds{1}(\cdot)$ is a indicator function.

For a demographic group $g$ of certain race, gender, or age, etc., we can further calculate its own FPR $\gamma^{+}_{g}$ and FNR $\gamma^{-}_{g}$ as follows:
\begin{equation}
\label{eq:false_positive_rate_group}
\gamma^{+}_{g}=\frac{\sum_{i=1}^{N^{-}_{g}} \mathds{1}({S}^{-}_{g}[i] > T_{u})}{N^{-}_{g}} ,
\end{equation}
\begin{equation}
\label{eq:false_negative_rate_group}
\gamma^{-}_{g}=\frac{\sum_{i=1}^{N^{+}_{g}} \mathds{1}({S}^{+}_{g}[i] < T_{u})}{N^{+}_{g}} ,
\end{equation}
where ${S}^{-}_{g}$, ${S}^{+}_{g}$, $N^{-}_{g}$ and $N^{+}_{g}$ are the corresponding numbers and similarity sets of the $g$ group.
To analyze bias in recognition performance, we adopt the FPR and FNR protocols to exhibit the performance difference across different demographic groups. 
We employ the BUPT-Balanced dataset~\cite{wang2020mitigating} to train a ResNet-34 model with ArcFace and show the FPR and FNR performance on RFW~\cite{wang2019racial} in Fig.~\ref{fig:demographic_difference_in_RFW}.
By comparing the FPR and FNR of four races, we notice that the performance of different races varies greatly, and moreover, the difference in FPR is much larger than that in FNR (the standard deviation of FPR at $T_{u}=0.31$ is $7.6$, while the standard deviation of FNR at $T_{u}=0.31$ is $0.97$).
We note that similar results are also reported in the NIST FRVT~\cite{ngan2015face}. 

Considering such results, we believe that achieving higher consistency in FPR prior to FNR across demographics is essential for improving fairness in face recognition. Besides, we define the standard deviation of the ratio between the demographic and overall FPR as bias degree. Given the group set $\mathcal{G}$ and group number $N_{\mathcal{G}}$, the bias degree $\delta$ is formulated as follows:
\begin{equation}
\label{eq:bais_degree}
\delta = \frac{1}{N_{\mathcal{G}}} 
         \sqrt{\sum_{g \in \mathcal{G}}\left(\frac{\gamma^{+}_{g} - \mu}{\gamma^{+}}\right)^2}
\end{equation}
where $\mu$ is the average demographic FPR. 
In our following experiments, this criterion is used as a fairness evaluation protocol on several benchmarks.

\paragraph{Consistency of Instance FPR}
As an extreme case, if a demographic group is consisted of one single instance, demographic group's FPR degrades into instance's FPR. 
Correspondingly, the consistency of FPRs across different demographic groups is generalized as the consistency of FPRs across different instance. During training, we choose to increase such a generalized version of consistency rather than the consistency of demographic groups' FPRs to mitigate face recognition bias. There are two reasons: 1) Demographic groups can be divided by various kinds of attributes, such as race, gender, and age etc., which is too numerous to enumerate; 2) Evaluating the consistency of instance FPR requires no demographic annotations of image samples. 
In the following, we first revisit the softmax-based losses and then show our way to achieve a higher FPR consistency via introducing extra false positive penalties into the softmax-like loss.

\begin{figure}[t]
  \centering
  \includegraphics[trim={0 0 0 0mm},clip,width=1\linewidth]{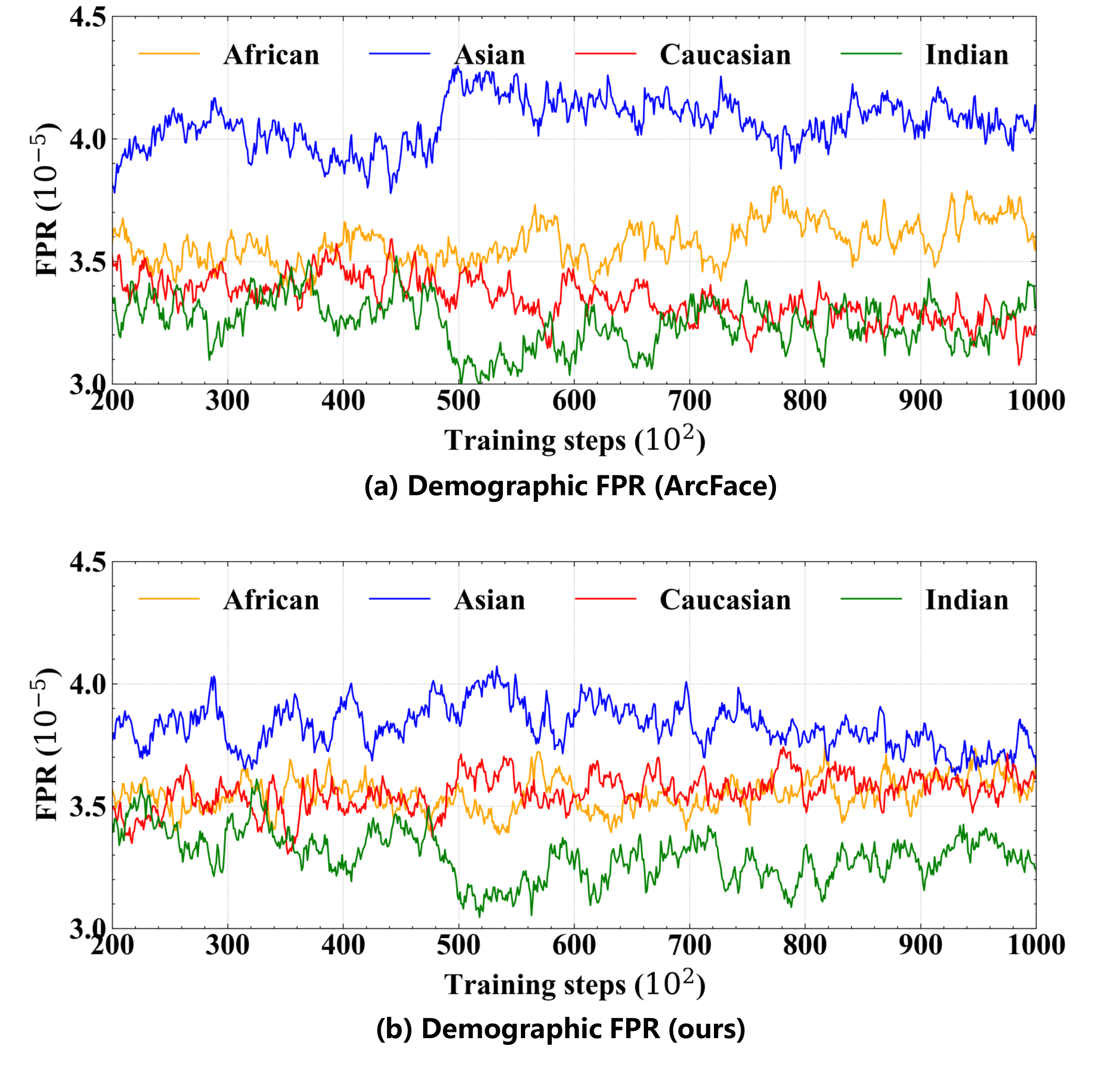}
  \caption{\small Comparison on demographic FPR in training.
   We compare the FPR trends of different races with ArcFace and our methods, respectively. 
   In ArcFace (a), the FPRs of Caucasian and Indian are still at a low level, while the FPRs of black and Asian are at a high level, even continue to grow.
   In our method (b), except Indian, the FPRs of the other races trend to converge at the end of training.
   We notice that the FPR of Caucasian is a slightly increased. However, we evaluate this trained model on RFW, and the performance of Caucasian is not degraded.}
  \label{fig:statistics_on_FPR_demographic_groups_in_training}
\end{figure}

\begin{figure}[t]
  \centering
  \includegraphics[trim={0 0 0 0mm},clip,width=1\linewidth]{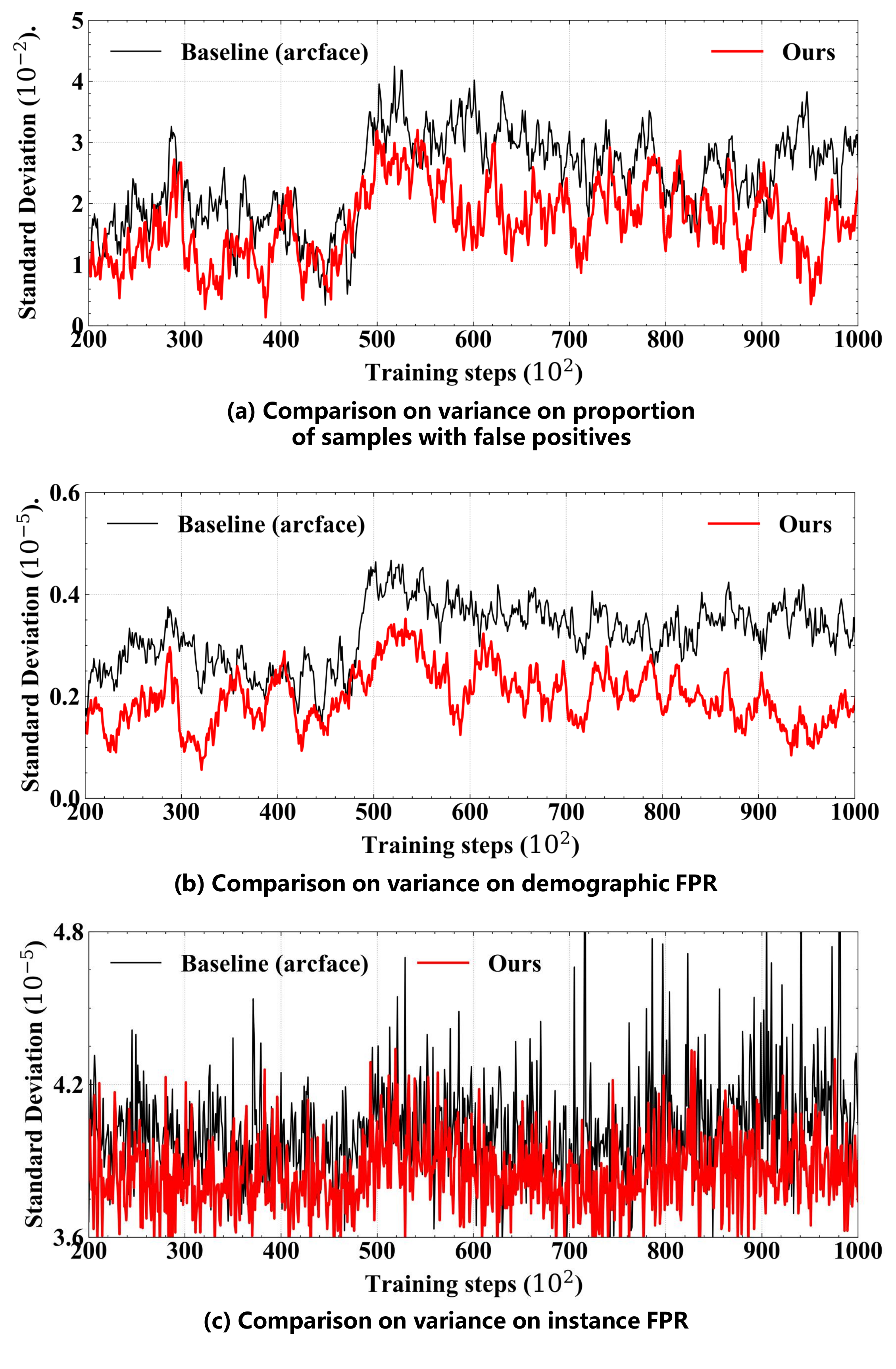}
  \caption{\small Comparison on variance in training. 
  We compare the training variance in three aspects.
  As shown in (a), in a mini-batch, the proportion of the samples whose instance FPR is greater than 0 is compared between baseline and our method.
  (b) shows the standard deviations on demographics FPR, and (c) shows the standard deviations on instance FPR. Compared with arcface, our method achieves higher consistency on demographic FPR and instance FPR.}
  \label{fig:statistics_on_var_demographic_groups_in_training}
\end{figure}

\subsection{FPR Penalty Loss}

\paragraph{Softmax Loss Function.}
The original softmax loss is formulated as follows:
\begin{equation}
\label{eq:softmax}
{
\mathcal{L} = -\log\frac{e^{W_{y_{i}}x_{i}+b_{y_{i}}}}{\sum^{n}_{j=1}e^{W_{j}x_{i}+b_{j}}},
}
\end{equation}



where $x_{i}\in R^d$ denotes the deep feature of $i$-th sample which belongs to the $y_{i}$ class, $W_{j} \in R^d$ denotes the $j$-th column of the weight $W\in R^{d\times n}$ and $b_j$ is the bias term.
The class number and the embedding feature size are $n$ and $d$, respectively.
In practice, the bias is usually set to $b_{j} = 0$ and the individual weight is set to $||W_{j}||=1$ by $l_2$ normalization. The deep feature is also normalized and re-scaled to $s$. Thus, the original softmax can be modified as follows:
\begin{equation}
\label{eq:softmax_modify}
{
\mathcal{L} = -\log\frac{e^{s(\cos\theta_{y_{i}})}}{e^{s(\cos\theta_{y_{i}})}+
                \sum^{n}_{j\neq y_{i}}e^{s(\cos\theta_{j})}}.
}
\end{equation}
Since the learned features with the original softmax loss may not be discriminative enough for practical face recognition problem, several variants are proposed and can be formulated in a general form:
\begin{equation}
\label{eq:margin_based_softmax_first}
{
\mathcal{L} = 
-\log\frac{e^{s\cdot G(\cos{\theta_{y_i}})}}
          {e^{s\cdot G(\cos{\theta_{y_i}})} +
           \sum^{n}_{j\neq y_{i}} e^{s\cdot H(\cos{\theta_{j}})}},
}
\end{equation}
where $G(\cos{\theta_{y_i}})$ and $H(\cos{\theta_{j}})$ are the functions to modulate the positive and negative cosine similarities, respectively.
In margin-based loss function, such as ArcFace, $G(\cos{\theta_{y_i}})=cos{(\theta_{y_i} + m)}$ aims to emphasize the inter-class similarity. In mining-based loss functions, $H(\cos{\theta_{j}})$ is designed to mining difficult negative pairs to decrease the intra-class confusion. 
However, the previous works focus on improving the discrimination performance on popular benchmarks, but not on enhancing the fairness of performance. 
Next, we introduce an extra false positive penalty term into the softmax-based losses, aiming at making the instance FPR more consistent and consequently enhancing the fairness of face recognition performance.

\paragraph{Extra Penalty on the FPR of Instance.} 
Since the $y_i$-th column of the weight $W$ usually could be regarded as a representative of the $y_i$-th class, for the $i$-th instance belonging to class $y_i$, the target logit $\cos{\theta_{y_i}}$ could be considered as the similarity of a positive pair, while the non-target logits $\cos{\theta_{j}}$, ${j\neq y_{i}}$ could be considered as the similarities of negative pairs.
According to Eq.~\ref{eq:false_positive_rate_group}, given these non-target similarities and a unified threshold $T_{u}$, corresponding to a overall FPR $\gamma^{+}_{u}$,  the FPR of the instance can be calculated as:
\begin{equation}
\label{eq:false_positive_rate_instance_non_target_logit}
\gamma^{+}_{i}=\frac{\sum_{j=1,j\neq y_{i}}^{n} \mathds{1}(\cos{\theta_{j}} > T_{u})}{n-1},
\end{equation}

To make instance FPRs more consistent, that is, all close to FPR $\gamma^{+}_{u}$, we add an extra penalty term in the denominator of the softmax function which is in proportion to the ratio of instance FPR to overall FPR $\gamma^{+}_{i}/\gamma^{+}_{u}$. Specially, we add the ratio (multiplied by a factor $\alpha>0$) to the original non-target logit, leading to the loss function presented as follows: 

\begin{equation}
\label{eq:our_proposed_loss}
{
\mathcal{L} = 
-\log\frac{e^{s\cdot G(\cos{\theta_{y_i}})}}
          {e^{s\cdot G(\cos{\theta_{y_i}})} +
           \sum^{n}_{j\neq y_{i}} e^{s\cdot \left(\cos{\theta_{j}}+\alpha\frac{\gamma^+_{i}}{\gamma^+_{u}}\right)}}.
}
\end{equation}

Since the extra penalty ${e^{s\alpha \gamma^+_{i}/\gamma^+_{u}}}$ is always $>1$, a larger instance-overall FPR ratio yields a larger loss value. By such unequal penalties, the instance FPRs are supposed to be much consistent. Further, considering the fact that more attention should be paid to those false positive cases with higher similarity (hard samples), we introduce a weighted FPR function of instance as follows:
\begin{equation}
\label{eq:false_positive_rate_instance_panelty}
\Bar{\gamma}^{+}_{i}=\frac{\sum_{j=1,j\neq y_{i}}^{n} \mathds{1}(\cos{\theta_{j}} > T_{u})
                \cdot F\left(\cos{\theta_{j}}\right)}{n-1}.
\end{equation}
Here the function $F(z)$ is supposed to give larger weights to false positive cases with higher similarities and thus should be monotone increasing. Without loss of generality, in this paper, we use the power function $F(z) = sgn\left(z \right) \left| z \right| ^{p}$ as the weighted function, where $p \geq 1$ and $sgn(\cdot)$ is the sign function. Since $T_{u}$ is usually positive, the sign function and the abs. function can be omitted, leading to $F(z) = z^{p}$. When $p=1$, $F(\cdot)$ degrades into $cos{\theta_{j}}$.

Finally, we show the effect achieved by the loss function on the training set. Fig.~\ref{fig:statistics_on_FPR_demographic_groups_in_training} shows the difference of the FPR trends between baseline and our method. 
In Fig.~\ref{fig:statistics_on_FPR_demographic_groups_in_training} (b), We notice that FPR of Asian in our method keeps decreasing in most time of training process, and becomes much consistent with other races at the end. 
In Fig.~\ref{fig:statistics_on_var_demographic_groups_in_training}, we compare the training variance on proportion of samples with false positives, demographic FPR and instance FPR.
As a result, both the variance in these three aspects are lower than the baseline method.
In a word, our algorithm helps to mitigate the race bias effectively.

\paragraph{FPR Setting and Threshold Estimation in Training.} 
From Eq.~\ref{eq:false_positive_rate_instance_non_target_logit} and Eq.~\ref{eq:false_positive_rate_instance_panelty}, we see that our proposed method relies on the choice of a overall $\gamma^{+}_{u}$, which further  involves the estimation of the threshold $T_{u}$. In practice, the choice of the FPR depends on the deployment scenario of the face recognition system. For example, to balance the risk and user experience, the FPR is usually set to $1e$-$5$ in a  face access control system. And the popular public face benchmarks often focus on the FPR range [$1e$-$1$, $1e$-$6$]. Note that a lower FPR means less number of false positive cases. If the overall FPR is set to be extremely small, there are rare  false positive cases can get extra penalty, which may intuitively lower the performance gain of our method. Besides, it's hard to estimate  a stable threshold corresponding to an extremely small FPR, because of a lack of enough negative pairs during training. For the above reasons, we choose the overall FPR range as [$1e$-$1$, $1e$-$5$] in this paper.

Given an overall FPR, usually the threshold is estimated from the quantile of the distribution of all negative pairs. Here, we utilize the non-target logits instead of the negative instance pairs in threshold estimation, for reasons that compared with the size of mini-batch, the number of class in training is often much larger, leading to more negative pairs and thus a more accurate and stable threshold estimate. 

\begin{algorithm}[t]
\small
\SetAlgoLined
\KwIn{The deep feature of $i$-th sample  with its label $y_{i}$, cosine similarity $\cos{\theta_{j}}$ of two vectors, last fully-connected layer parameters $W$, embedding network parameters $\Theta$, class number $c$, sample number $n$, learning rate $\lambda$, and overall false positive rate $\gamma^{+}_{u}$}
iteration  number $k\leftarrow 0$, parameter $t\leftarrow 0$, $\gamma^{+}_{u}\leftarrow 1e^{-4}$\;
 \While{not converged}{
  Compute the $\lceil \gamma^{+}_{u} n(c-1) \rceil$-th largest value of set $\left\{ \cos{\theta_{j}} \mid i\in [1,n], j\in [1,c], j\neq y_{i} \right\}$ as the temporary threshold $T_{u}$\;
  \eIf{$\cos(\theta_{j}) > T_{u}$}{
   $I_{j} = 1$\;
   }{
   $I_{j} = 0$\;
  }
  Compute the weighted FPR $\Bar{\gamma}^{+}_{i}$ by Eq.~\ref{eq:false_positive_rate_instance_panelty}\;
  Compute the loss $\mathcal{L}$ by Eq.~\ref{eq:our_proposed_loss} (replace ${\gamma}^{+}_{i}$ by $\Bar{\gamma}^{+}_{i}$)\;
  Compute the gradient of $W_{j}$ and $x_{i}$ by Eq.~\ref{eq:ours_gradient_theta}\;
  Update the parameters $W$ and $\Theta$ by:
  $W^{(k+1)} = W^{(k)} - \lambda^{(k)}\frac{\partial \mathcal{L}_{i}}{\partial W}$,
  $\Theta^{(k+1)} = \Theta^{(k)} - \lambda^{(k)}\frac{\partial \mathcal{L}_{i}}{\partial x_i}\frac{\partial x_i}{\partial \Theta^{(k)}}$\;
  $k \leftarrow k+1$\;
 }
\KwOut{$W$, $\Theta$.}
 \caption{FPR Penalty Loss}
 \label{alg:training}
\end{algorithm}

\paragraph{Optimization.}
We show our method~\ref{alg:training} can be easily optimized by the conventional stochastic gradient descent. Let's denote $G_{i}$ as the $sG(\cos{\theta_{y_{i}}})$ of the sample belongs to the $y_{i}$-th class, $H_{j}$ as $s\left(\cos{\theta_{j}}+\alpha{\gamma^+_{i}}/{\gamma^+_{u}}\right)$, and $I_{j}$ as the mining mask. In this section, we consider the CosFace form of our loss function, so $G_{i}=\cos{\theta_{y_{i}}}-m
$ and $H_{i}=\cos{\theta_{j}}+\alpha \Bar{\gamma}^{+}_{i}/\gamma^{+}_{u}$. In the backward propagation process, the gradients \textit{w.r.t.} $x_{i}$ and $W_{j}$ are presented as follows:
\begin{equation}
\small
\label{eq:ours_gradient_theta}
\begin{aligned}
&\frac{\partial \mathcal{L}_{i}}{\partial W_{y_{i}}} = 
    \frac{\partial \mathcal{L}_{i}}{\partial G_{i}} \cdot x_{i}, \\
&\frac{\partial \mathcal{L}_{i}}{\partial W_{j}} =
    \left(1 + \frac{\alpha}{\gamma^{+}_{u}}  \cdot
    I_{j} \frac{\partial F}{\partial \cos{\theta_{j}}} \right)  \cdot \frac{\partial \mathcal{L}_{i}}{\partial H_{j}} \cdot x_{i}, \\
&\frac{\partial \mathcal{L}_{i}}{\partial x_{i}} = 
    \frac{\partial \mathcal{L}_{i}}{\partial G_{i}} \cdot W_{y_{i}} + 
    \left(1 + \frac{\alpha}{\gamma^{+}_{u}} \cdot
    \sum_{j\neq y_{i}} I_{j} \frac{\partial F}{\partial \cos{\theta_{j}}} \right) \cdot \frac{\partial \mathcal{L}_{i}}{\partial H_{j}}\cdot W_{j},
\end{aligned}
\end{equation}
Based on the above formulations, we can find the extra gradients for alleviating bias have a new composed term $\small \frac{\partial \mathcal{L}_{i}}{\partial H_{j}}\frac{\partial F}{\partial \cos{\theta_{j}}}$, if the $j$-th logit with $I_{j}=1$ is chosen as a false positive case. The term $\small \frac{\partial \mathcal{L}_{i}}{\partial H_{j}}$ brings gradient adjustment from false positive cases above the threshold, while the term $\small\frac{\partial F}{\partial \cos{\theta_{j}}}$ further modulates the former adjustment by the similarity of specific false positive case.  
%

\section{Experiments}
\subsection{Experimental Setting}
\paragraph{Dataset.}
In this study, we employ BUPT-Balancedface and BUPT-Globalface dataset~\cite{wang2019racial} for training.
BUPT-Balancedface dataset contains $1.3$M images of $28$K celebrities and is approximately race-balanced with $7$K identities per race.
BUPT-Globalface dataset contains $2$M images of $38$K celebrities, and its racial distribution is approximately the same as the real distribution of the world’s population. 
RFW dataset~\cite{wang2020mitigating} and BFW dataset~\cite{robinson2020face} are used for fairness testing. 
RFW consists of faces from four race groups: African, Asian, Caucasian, and Indian.
Each race group contains nearly $10$K images of $3$K individuals for face verification. Compared with  RFW, the BFW dataset provides balanced face data with more attributes, including ID, gender, and race. There are eight demographic groups according to two genders and four ethnic groups (\textit{i.e.} Black, White, Asian, and  Indian), and each demographic group consists of $200$ subjects with $2.5$K images.

\paragraph{Training Setting.}
We follow~\cite{wang2018cosface, deng2019arcface} to crop the $112\times 112$ faces with five landmarks detected by MTCNN~\cite{zhang2016mtcnn}. 
The RGB images are first normalized by subtracting $127.5$ and divided by $128$, then feeding into the embedding network.
we adopt ResNet$34$, ResNet$50$ and ResNet$100$ as in~\cite{he2016deep, deng2019arcface} as the embedding network.
We conducted all the experiments on $8$ NVIDIA Tesla V$100$ GPU with Pytorch~\cite{paszke2017automatic} framework. 
The models are trained with SGD algorithm, with momentum $0.9$ and weight decay $5e-4$. 
The batch size is set to be $512$.
On BUPT-Balancedface, the learning rate starts from $0.1$ and is divided by $10$ at $20$, $32$, $36$ epochs.
The training process is finished at $40$ epochs.
On BUPT-Globalface, we divide the learning rate at $10$, $18$, $22$ epochs and finish at $24$ epochs. 
We follow the common setting as~\cite{wang2018cosface} to set $s=64$ and $m=0.35$.

\subsection{Ablation Study}
\paragraph{Effect of the overall FPR $\gamma_{u}^{+}$.}
We conduct experiments at five fixed FPRs from $10^{-5}$ to $10^{-1}$, and find that nearly all the best performance of training is achieved when $\gamma_{u}^{+}=10^{-4}$, except that a higher accuracy of Caucasian is obtained at $\gamma_{u}^{+}=10^{-5}$, as shown in Tab.~\ref{tab:ablation_study_different_fpr}.
We explain the reasons as follows:
1) When $\gamma_{u}^{+}$ is set to be $10^{-5}$ or even a lower value, a relatively large value of threshold is used to measure the FPR of instance and generate penalty terms. Correspondingly, the number of extra penalty would be reduced. Besides, considering that the noisy data (\textit{e.g.} label flips) is ubiquitous in training dataset, with a small number of noisy false positive cases, the accuracy of estimated threshold and the extra gradient adjustment may be affected dramatically.
2) When $\gamma_{u}^{+}$ is set to be a large value, \textit{e.g.} $10^{-1}$ or higher, we obtain a relatively small value of threshold. With such a threshold, most training instances would be forced to generate numerous false negative pairs, since the similarity of most negative pairs is around 0. As a result, the instance FPRs and its corresponding penalty would be almost equal and hard to be more consistent through optimization. 
Therefore, $\gamma_{u}^{+}=10^{-4}$ is a reasonable choice. We will use this configure in our following experiments.

\begin{table}[t!]
\begin{center}
\scriptsize
\caption{\small \textbf{Verification performance (\%) of different FPR parameter $\gamma$.}}
\label{tab:ablation_study_different_fpr}
\resizebox{1\columnwidth}{!}{
\begin{tabular}{l|cccc|c|c}
\hline
Methods (\%)                      & African & Asian & Caucasian & Indian  & Avg     & Std\\ \hline\hline
$\gamma_{u}^{+}=10^{-5}$                 & $95.60$ & $95.10$ & $\bf{97.18}$ & $96.32$ & $96.05$ & $0.91$ \\
$\gamma_{u}^{+}=10^{-4}$           & $\bf{95.95}$ & $\bf{95.17}$ & $96.78$ & $\bf{96.38}$ & $\bf{96.07}$ & $\bf{0.69}$ \\
$\gamma_{u}^{+}=10^{-3}$                 & $95.47$ & $94.90$ & $96.92$ & $96.12$ & $95.84$ & $0.87$ \\ 
$\gamma_{u}^{+}=10^{-2}$                 & $95.45$ & $94.78$ & $96.98$ & $96.13$ & $95.84$ & $0.94$ \\
$\gamma_{u}^{+}=10^{-1}$                 & $95.23$ & $94.60$ & $95.87$ & $95.97$ & $95.42$ & $0.64$ \\
\hline
\end{tabular}
}
\end{center}
\end{table}

\begin{table}[t!]
\begin{center}
\scriptsize
\caption{\small \textbf{Verification performance (\%) of different exponent $p$ in $F(z)$.}}
\label{tab:ablation_study_different_n}
\resizebox{1\columnwidth}{!}{
\begin{tabular}{l|cccc|c|c}
\hline
Methods (\%)      & African & Asian & Caucasian & Indian  & Avg     & Std\\ \hline\hline
$p=0.25$          & $95.35$ & $95.10$ & $96.97$ & $96.07$ & $95.87$ & $0.84$ \\
$p=0.5$           & $95.27$ & $94.93$ & $96.58$ & $96.02$ & $95.70$ & $0.74$ \\
$p=1.0$           & $95.18$ & $94.92$ & $96.90$ & $95.83$ & $95.71$ & $0.88$ \\
$p=1.5$           & $95.27$ & $94.67$ & $97.05$ & $96.23$ & $95.80$ & $1.05$ \\ 
$p=2.0$           & $\bf{95.95}$ & $95.17$ & $96.78$ & $\bf{96.38}$ & $\bf{96.07}$ & $\bf{0.69}$ \\
$p=2.5$           & $95.85$ & $95.00$ & $96.96$ & $96.20$ & $96.00$ & $0.82$ \\
$p=3.0$           & $95.60$ & $\bf{95.18}$ & $\bf{97.17}$ & $95.98$ & $95.98$ & $0.85$ \\
\hline
\end{tabular}
}
\end{center}
\end{table}

\paragraph{Effect of exponent $p$ in $F(z)$.}
With the fixed FPR as $10^{-4}$ , we further investigate the effect of exponent $p$ in $F(z) =  z^{p}$. Here, we set the value varies from 0.25 and 3.0. 
Tab.~\ref{tab:ablation_study_different_n} shows that the performance of our method decreases as $n$ increasing from 0.25 to 1.0, and then gradually increases until n reaches near 2.0, and then the performance begins to decrease again.
Note that when $p>1$, the gradient is convex with regard to $z$, and with a moderate value of $p$, e.g.$p=2$, we can give a proper but not too much penalty on false positive cases with larger similarity. Based on these reasons, we choose $p=2$ in our following experiments.


\begin{table}[t!]
\begin{center}
\scriptsize
\caption{\small \textbf{Verification performance (\%) of protocol on RFW with SOTA methods} ([BUPT-Balancedface]).}
\label{tab:rfw_buptbalanced}
\resizebox{1\columnwidth}{!}{
\begin{tabular}{l|cccc|c|c}
\hline
Methods (\%)                      & African & Asian & Caucasian & Indian  & Avg     & Std\\ \hline\hline
ArcFace-R34 \cite{wang2020mitigating} & $93.98$ & $93.72$ & $96.18$ & $94.67$ & $94.64$ & $1.11$ \\
CosFace-R34 \cite{wang2020mitigating} & $92.93$ & $92.98$ & $95.12$ & $93.93$ & $93.74$ & $1.03$ \\
DebFace-R34 (ECCV'$20$)               & $93.67$ & $94.33$ & $95.95$ & $94.78$ & $94.68$ & $0.83$ \\
PFE-R34 \cite{gong2020mitigating}     & $95.17$ & $94.27$ & $96.38$ & $94.60$ & $95.11$ & $0.93$ \\
GAC-R34 \cite{gong2020mitigating}     & $94.65$ & $94.93$ & $96.23$ & $95.12$ & $95.23$ & $0.60$ \\
RL-RBN-R34(cos) (CVPR'$20$)           & $95.27$ & $94.52$ & $95.47$ & $95.15$ & $95.10$ & $\bf{0.41}$ \\
RL-RBN-R34(arc) (CVPR'$20$)           & $95.00$ & $94.82$ & $96.27$ & $94.68$ & $95.19$ & $0.93$ \\
\textbf{Ours}-R34  & $\bf{95.95}$ & $\bf{95.17}$ & $\bf{96.78}$  & $\bf{96.38}$ & $\bf{96.07}$ & ${0.69}$ \\
\hline\hline
ArcFace-R50     & ${95.55}$     & ${94.95}$    & ${96.68}$     & ${95.47}$     & ${95.66}$     & ${0.73}$ \\
\textbf{Ours}-R50   & $\bf{96.47}$  & $\bf{95.75}$ & $\bf{97.08}$  & $\bf{96.77}$  & $\bf{96.52}$  & $\bf{0.57}$ \\
\hline\hline
ArcFace-R100     & ${96.43}$     & ${94.98}$    & ${97.37}$     & ${96.17}$     & ${96.24}$     & ${0.98}$ \\
\textbf{Ours}-R100   & $\bf{97.03}$  & $\bf{95.65}$ & $\bf{97.6}$  & $\bf{96.82}$  & $\bf{96.78}$  & $\bf{0.82}$ \\
\hline
\end{tabular}
}
\end{center}
\end{table}

\begin{figure*}[t!]
  \centering
  \includegraphics[trim={0 0 0 0mm},clip,width=0.98\linewidth]{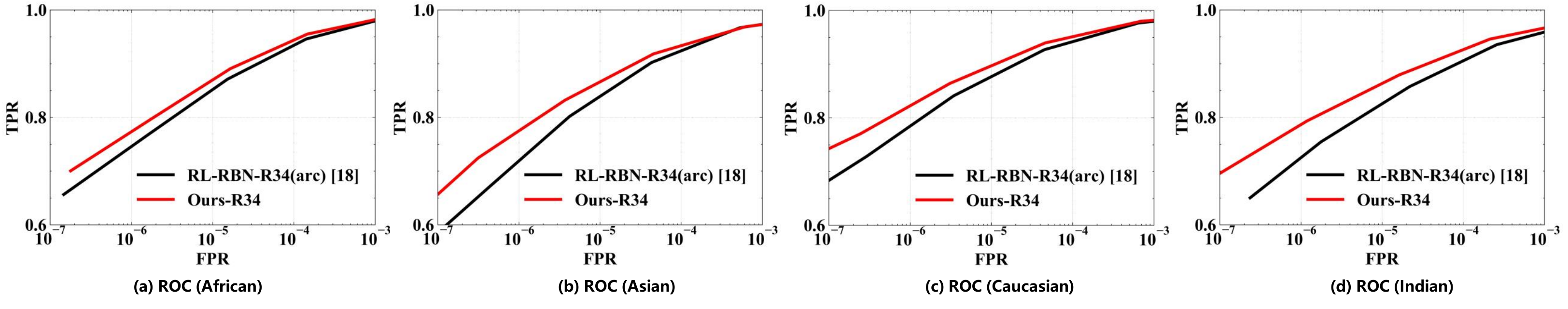}
  \caption{\small ROC for RFW.} 
  \label{fig:ROC_for_RFW}
\end{figure*}

\begin{figure*}[t!]
  \centering
  \includegraphics[trim={0 0 0 0mm},clip,width=0.98\linewidth]{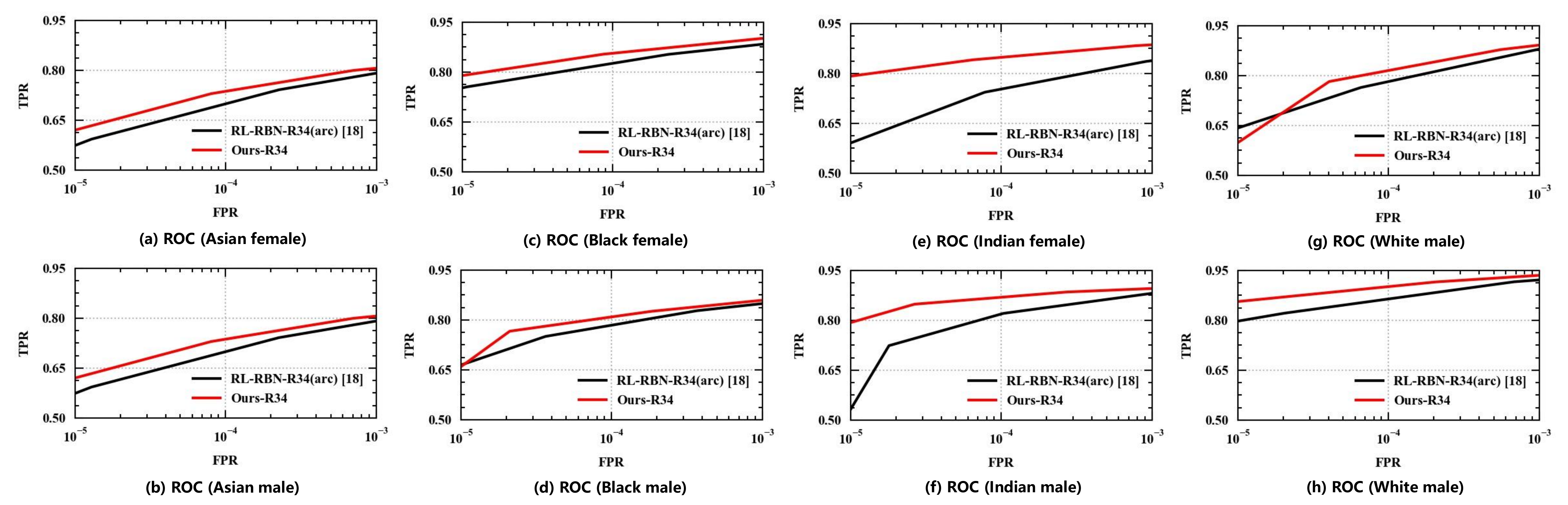}
  \caption{\small ROC for BFW.} 
  \label{fig:ROC_for_BFW}
\end{figure*}

\begin{table}[t!]
\begin{center}
\scriptsize
\caption{\small \textbf{Verification accuracy (\%) of protocol on RFW with SOTA methods} ([BUPT-Globalface]).}
\label{tab:rfw_buptglobalface}
\resizebox{1\columnwidth}{!}{
\begin{tabular}{l|cccc|c|c}
\hline
Methods (\%)                      & African & Asian & Caucasian & Indian & Avg & Std\\ \hline\hline
ArcFace-R34 \cite{wang2020mitigating} & $93.87$ & $94.55$ & $97.37$ & $95.86$ & $95.37$ & $1.53$ \\
CosFace-R34 \cite{wang2020mitigating} & $92.17$ & $93.50$ & $96.63$ & $94.68$ & $94.25$ & $1.90$ \\
RL-RBN-R34(cos) (CVPR'$20$)           & $94.27$ & $94.58$ & $96.03$ & $95.15$ & $95.01$ & $0.77$ \\
RL-RBN-R34(arc) (CVPR'$20$)           & $94.87$ & $95.57$ & $97.08$ & $95.63$ & $95.79$ & $0.93$ \\
\textbf{Ours}-R34  & $\bf{95.77}$ & $\bf{95.85}$ & $\bf{97.92}$  & $\bf{96.70}$ & $\bf{96.56}$ & $\bf{0.75}$ \\
\hline\hline
ArcFace-R50     & ${96.23}$     & ${96.43}$    & ${97.98}$     & ${96.92}$     & ${96.89}$     & ${0.78}$ \\
\textbf{Ours}-R50   & $\bf{96.85}$  & $\bf{96.75}$ & $\bf{98.30}$  & $\bf{96.95}$  & $\bf{97.21}$  & $\bf{0.73}$ \\
\hline\hline
ArcFace-R100     & ${96.68}$     & ${96.10}$    & ${98.17}$     & ${97.32}$     & ${97.07}$     & ${0.89}$ \\
\textbf{Ours}-R100   & $\bf{97.37}$  & $\bf{96.48}$   & $\bf{98.57}$  & $\bf{97.4}$  & $\bf{97.45}$  & $\bf{0.85}$ \\
\hline
\end{tabular}
}
\end{center}
\end{table}

\begin{table}[t!]
\begin{center}
\scriptsize
\caption{\small \textbf{Bias degree of protocol on RFW with SOTA methods.}}
\label{tab:rfw_fpr_variance}
\vspace{2mm}
\resizebox{0.9\columnwidth}{!}{
\begin{tabular}{l|cccc}
\hline
overall FPR        & $10^{-5}$     & $10^{-4}$     & $10^{-3}$    & $10^{-2}$ \\\hline\hline
RL-RBN-R34(arc)    & $351.98$      & $208.44$      & $92.18$      & $16.70$ \\
\textbf{Ours}-R34  & $\bf{257.53}$ & $\bf{185.91}$ & $\bf{59.25}$ & $\bf{10.33}$
\\\hline
\end{tabular}
}
\end{center}
\end{table}

\begin{table}[t!]
\begin{center}
\scriptsize
\caption{\small \textbf{Bias degree of protocol on BFW with SOTA methods.}}
\label{tab:bfw_fpr_variance}
\vspace{2mm}
\resizebox{0.9\columnwidth}{!}{
\begin{tabular}{l|ccccc}
\hline
overall FPR         & $10^{-7}$     & $10^{-6}$     & $10^{-5}$     & $10^{-4}$    & $10^{-3}$ \\\hline\hline
RL-RBN-R34(arc)     & $2.44$        & $2.01$        & $2.49$        & $2.91$       & $2.43$       \\
\textbf{Ours}-R34   & $\bf{1.18}$   & $\bf{1.08}$   & $\bf{1.18}$   & $\bf{1.67}$  & $\bf{1.80}$ 
\\\hline
\end{tabular}
}
\end{center}
\end{table}

\subsection{Comparisons with SOTA methods}
\paragraph{Accuracy on RFW.}
We train a ResNet$34$ model on BUPT-Balancedface with our method, and report the results of the competitors following the RFW protocol, shown in Tab.~\ref{tab:rfw_buptbalanced}. 
Our method shows superiority over the competitors with the balanced dataset. 
Compared with the SOTA results, it achieves about $0.77\%$ gains for average accuracy, and its standard deviation is decreased to $0.69$ which is slightly higher than SOTA. 
Though the standard deviation of GAC and RL-RBN(cos) is much lower, its performance on Caucasian is actually worse than that of the CosFace baseline. In contrast, for our method, the reduction in bias is obtained along with the accuracy improvement of all four races.  
We also train a ResNet$34$ model on BUPT-Globalface with our method and arcface. In Tab.~\ref{tab:rfw_buptglobalface}, it shows that the average accuracy of our method is still much better than other competitors, while our standard deviation is also lower than others.
Besides, we train Arcface and our method with ResNet$50$ and ResNet$100$ as in~\cite{deng2019arcface}. 
As shown in Tab.~\ref{tab:rfw_buptbalanced} and Tab.~\ref{tab:rfw_buptglobalface}, our method also performs better than the common baseline.
The above results show that our method can achieve competitive performances on both race balanced and unbalanced datasets, with regard to the mean and standard deviation of accuracy.

\paragraph{FPR on RFW.}
According to the FPR and TPR evaluation protocols discussed in Sec.2, we compare the performance between baseline and our method.
Fig.~\ref{fig:ROC_for_RFW} (a) shows the African ROC curves of our method and the SOTA competitor, and it is clear that our method performs best. 
Besides, Fig.~\ref{fig:ROC_for_RFW} (b)(c)(d) respectively show the other groups' ROC curves.
This experiment on RFW proves that our loss leads to the face  recognition model with more discriminative features than RL-RBN(arc).
According to the evaluation protocol defined in Eq.~\ref{eq:bais_degree}, we also compare the bais degree with the SOTA method in Tab.~\ref{tab:rfw_fpr_variance}.
The lower bias degree at each threshold corresponding with the overall FPR demonstrates our method can achieve better performance on fairness recognition than that of RL-RBN(arc).

\paragraph{FPR on BFW.}
Fig.~\ref{fig:ROC_for_BFW} shows the ROC curves on all 8 demographic groups in BFW.
Across all ethnicity, our method achieves better performance on the female group and the male group.
Across all gender, the TPR on each ethnicity in our method is much better than RL-RBN(arc).
As defined in Eq.~\ref{eq:bais_degree}, we calculate the bias degree on BFW shown in Tab.~\ref{tab:bfw_fpr_variance}, which proves that our algorithm also can alleviate both gender and race bias across demographics effectively.

\section{Conclusions}
In this paper, we develop a novel penalty term into the softmax loss function to alleviate bias and improve the fairness performance in face recognition. We propose the concept of instance FPR as an extreme case of demographic FPR, and convert consistency of instance FPR as a penalty item of softmax-based loss. Extensive experiments on popular facial benchmarks demonstrate the effectiveness of our method compared to the SOTA competitors. Following the main idea of this work, future research can be expanded in various aspects, including designing a better weight function $F(\cdot)$ for inconsistency penalty, and investigating the effects of noise samples that might be mistakenly optimized as false positive cases.


{\small
\bibliographystyle{ieee_fullname}
\bibliography{paperbib}
}

\end{document}